\documentclass[11pt,a4paper]{article}

\usepackage[T1]{fontenc}
\usepackage[margin=1in]{geometry}
\usepackage{amsmath,amssymb}
\usepackage{graphicx}
\usepackage{booktabs}
\usepackage{tabularx}
\usepackage{url}
\usepackage[hidelinks]{hyperref}
\usepackage{authblk}
\usepackage{abstract}
\usepackage[round,authoryear]{natbib}   

\usepackage{titlesec}
\titleformat*{\section}{\large\bfseries}
\titleformat*{\subsection}{\normalsize\bfseries}

\title{\bfseries
       Institutions and the transmission of upper-tail human capital:
       scientific lineages across a millennium}

\author[1]{Hiroyuki Chuma\thanks{Corresponding author: chuma@iir.hit-u.ac.jp.
           Institute of Innovation Research, Hitotsubashi University,
           2-1 Naka, Kunitachi, Tokyo 186-8603, Japan.}}
\author[2]{Kanji Otsuka}
\author[3]{Yoichi Sato}

\affil[1]{Institute of Innovation Research, Hitotsubashi University}
\affil[2]{Professor Emeritus, Meisei University}
\affil[3]{Shuhari System}

\date{\today}

\begin{document}

\maketitle


\begin{abstract}\noindent
What made useful knowledge cumulative was not discovery alone but the
institutions that transmitted it. We provide the first exhaustive structural
measurement of the network through which upper-tail human capital passed from
master to student across a millennium. Using 470{,}000 mentor--student records
from Wikidata (which integrates the Mathematics Genealogy Project and MacTutor Archive), and all 64 historical Fields Medalists as a fixed, \emph{ex ante}
tracer set, backward traversal yields a directed acyclic graph of 25.5 million
paths reaching 57 generations. We document two institutional transitions.
First, a 17th-century watershed concentrates lineage traffic on Leibniz:
47 of 64 lineages pass through him with a 10:1 downstream-to-upstream ratio,
and seven independent attributes---learned-society membership (a 46-fold rise
per scholar), field, language, employer, institutional diversification, student
production, and diffusion entropy---re-organize coherently across the same
window. This is the network signature of Mokyr's Republic of Letters, and it
reframes the Newton--Leibniz priority dispute as a distinction between the
\emph{possession} and the \emph{transmission} of upper-tail human capital: it is
transmission that generates the spillovers on which growth depends. Second, 84\%
of lineages converge upstream on five 12th--13th-century Islamic and Byzantine
scholars before terminating at an 11th-century boundary---the ``Monastery
Wall''---at which personal academic mentorship first becomes record-generating
in Europe. Our claims are descriptive-structural, not causal. Because exhaustive
traversal at this scale defeats standard tools, we also contribute a
deterministic, algebraic graph-traversal instrument whose measurement bias we
characterize in closed form, and report one emergent property of independent
methodological interest.


\medskip
\noindent\textbf{Keywords:} Knowledge transmission; Human capital; Institutions;
Scientific networks; Cliometrics; Computational history.
\end{abstract}

\bigskip

\section{Introduction}

What enabled the sustained accumulation of the ``useful knowledge'' on which
modern economic growth depends? \citet{Mokyr2002,Mokyr2016} argues that the
answer lies not in any single discovery but in the institutions---universities,
learned societies, journals, correspondence networks---that made knowledge
transmittable across generations and borders, with the 17th--18th-century
Republic of Letters as the critical innovation. A complementary literature
locates the engine of growth specifically in the upper tail of the human-capital
distribution: \citet{Squicciarini2015} show that it is the density of
knowledge elites, not average literacy, that predicts industrial growth in the
age of Enlightenment. Yet the object common to both arguments---the network
through which upper-tail human capital actually moved from master to student,
across civilizations and across the porous boundaries between mathematics,
natural philosophy, and astronomy that characterized the pre-modern learned
world---has remained essentially unmeasured at scale.

This paper measures it. Our contribution is, first, a body of new structural
facts about that network, and second, a measurement strategy that makes such
facts obtainable. Existing empirical work on the economic history of knowledge
has measured narrow windows with specific proxies: \citet{Squicciarini2015}
on \emph{Encyclop\'edie} subscriptions, \citet{Dittmar2011} on the printing
press, \citet{Cantoni2014} on medieval universities. These designs achieve
causal identification within a bounded setting. The aggregate transmission
network---how many routes connected past to present across a full millennium,
where they converged, where they broke---calls for a different and
complementary mode: exhaustive rather than sampled, structural rather than
episodic, spanning a millennium rather than a decade. We are explicit at the
outset that the resulting facts are descriptive-structural, not causal
estimates. We regard this as the appropriate scope for a first map of an
object this large, and as a precondition for---not a substitute
for---identified work that would treat transmission costs as endogenous.

We document two transitions in the social architecture of transmission. The
first, at the 17th-century Leibnizian watershed, is a re-organization of
knowledge production within an already-existing institutional regime. The
second, at the 11th century, is a more foundational shift in the medium of
transmission---from communal monastic education to personal academic
lineage---which we name the \emph{Monastery Wall} after the network-legible
signature it leaves. Each is a structural property of a 25.5-million-path
network we reconstruct for the first time.

We use the complete set of historical Fields Medalists (64 laureates) solely
as a tracer set: a uniform, \emph{ex ante} criterion for which present-day
research lineages to follow backward through the mentor--student relation. The
Medal is not a claim about the primacy of mathematics; in Wikidata, which
integrates the Mathematics Genealogy Project (P549) and the MacTutor Archive
(P1563), the field label is a coarse modern tag that loses disciplinary purity
upstream. The point of a fixed starting set is replicability: any reader can
recover which nodes seed the 25.5 million paths, while the findings concern the
shape of the entire induced subgraph. A different tracer set (say, Nobel
laureates in physics) would change absolute counts but not the question the
exercise makes well posed: where, on a graph too large for unaided historical
intuition, does the topology concentrate?

A measurement obstacle had to be cleared first. Academic genealogy on
Wikidata's 470{,}000-record graph is a directed acyclic graph whose distinct
paths grow multiplicatively with depth, reaching 25.5 million at generation 57;
the standard lookup primitives of computational social science exhaust memory
far below this frontier size\footnote{In our PyVaCoAl application, the Frontier Size (FS) bounds the number of concurrent paths tracked during retrieval, preventing combinatorial explosion during branching reasoning. }, and standard graph centralities are semantically
blind to field. We therefore built a deterministic, algebraic graph-traversal
instrument (VaCoAl/PyVaCoAl). Because building the instrument was integral to
the measurement, we present it in the body (Section~\ref{sec:robust}) rather
than as a black box: we report its architecture, its biases, and one emergent
property---an instrument-level path-dependence that, on reflection, sharpens
rather than confounds the historical reading---and give the full algorithmic
specification in the Appendix. A reader interested only in the historical
findings may take from Section~\ref{sec:robust} the single fact that the
instrument's measurement error is bounded in closed form and is overwhelmed by
the first-order results.


\section{The structure of a millennium of transmission}

\subsection{The 17th-century watershed: an hourglass at Leibniz}

\begin{figure}[!htbp]
  \centering
  \includegraphics[width=\linewidth]{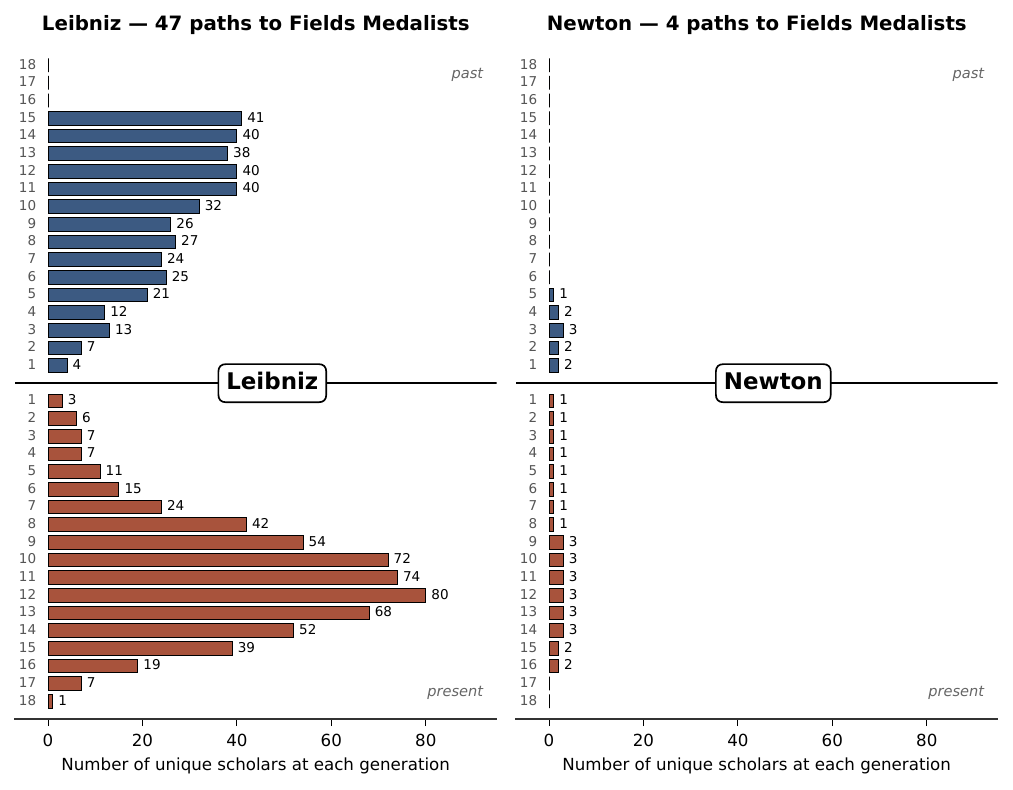}
  \caption{\textbf{Hourglass topology of genealogical traffic.}
    Leibniz~(top) versus Newton~(bottom), drawn to the same scale. Each bar is
    the number of unique scholars at a given generational distance through which
    paths to Fields Medalists pass. Upper half: mentor direction (past); lower
    half: student direction (present). The number to the left of each bar is the
    generational distance; to the right, the hub count. Leibniz exhibits the
    canonical hourglass; Newton's profile is a thin thread with no hub structure
    in either direction.}
  \label{fig:hourglass}
\end{figure}

Genealogical traffic centered on Leibniz forms a canonical hourglass
(Fig.~\ref{fig:hourglass}): a thin band of mid-stream positions concentrates
traffic while wide sets of more distant nodes feed in from the past and diffuse
out toward the present. A scholar who both draws from a diverse upstream pool
and scatters that inheritance across a diverse downstream pool acts as a
re-publication device for the entire graph; the hourglass is the macroscopic
signature of a teacher whose intellectual descendants remain visible for
centuries.

Upstream nodes carry an average of 5.3 traversing paths; downstream, the figure
expands to 53.4---a thickness ratio of $10.1\times$ (Table~\ref{tab:hubs}).
Newton's profile is the stark contrast: only four Fields Medalists trace their
genealogy through Newton, and his node shows no constriction-and-explosion
pattern in either direction. This is not a valuation of ``German networking''
over ``English genius''; it is a property of a directed graph built from a
single predicate, and the appropriate comparison is between two figures who
both carry enormous semantic weight in the history of the calculus yet differ
sharply in the depth and width of the lineages documented through them.

What Leibniz built was an apparatus of transmission: he founded the Berlin
Academy of Sciences in 1700, maintained one of the largest scholarly
correspondence networks in Europe, and initiated an academic lineage running
through Johann Bernoulli, Euler, Lagrange, Laplace, and Poisson. Newton, for
all his genius, operated largely in isolation and trained few successors. Other
hubs (Euler, Gauss, Hilbert; Table~\ref{tab:hubs}) also exhibit high thickness
ratios, but Leibniz combines a large absolute path count with the $10\!:\!1$
asymmetry, reflecting the scale of the lineage that actually passes through him.

The 300-year priority dispute over the calculus \citep{Hall1980} turned on a
question---who discovered first?---to which our instrument can add no evidence.
It can, however, reframe the question in economic terms. Rather than asking who
possessed the intellectual content first, we ask who built the transmission
infrastructure. We construct a composite \emph{Giant Score}
\begin{equation}
G(v) \;=\; \hat{s}(v)\cdot \hat{t}(v),
\qquad
\hat{s}(v) = \max\!\bigl(s(v)-0.5,\,0\bigr)\times 10,
\label{eq:giantscore}
\end{equation}
where $\hat{s}(v)$ is a measure of the node's calculus-semantic content and
$\hat{t}(v)$ is its path count to the 64 Fields Medalists, normalized by the
maximum (construction in Appendix~D). Only scholars high on
\emph{both} dimensions rank at the top (Table~\ref{tab:giant}). Newton ranks
91st despite a calculus-content score of $0.534$, almost indistinguishable from
Leibniz's $0.569$; his low rank is driven entirely by a path count of four.
Read through the upper-tail human-capital framework of
\citet{Squicciarini2015}, the table operationalizes a distinction long latent
in the history of science: between the \emph{possession} and the
\emph{transmission} of upper-tail human capital. Newton possessed the calculus;
Leibniz possessed \emph{and} transmitted it. It is transmission, not
possession, that generates the human-capital spillovers on which sustained
growth depends, and reframed this way the dispute admits a quantitative answer
the priority frame by construction cannot offer.

\begin{table}[!htbp]
  \centering
  \caption{Path counts and thickness ratios at hub nodes.}
  \label{tab:hubs}
  \begin{tabular}{lrrr}
    \toprule
    Hub      & Mentor avg.\ paths & Student avg.\ paths & Ratio \\
    \midrule
    Leibniz  &  5.3 & 53.4 & $10.1\times$ \\
    Gauss    &  2.7 & 41.2 & $15.1\times$ \\
    Hilbert  &  1.6 & 29.2 & $18.5\times$ \\
    Euler    &  3.6 & 52.4 & $14.6\times$ \\
    Newton   &  $\sim$1 & $\sim$2 & --- \\
    \bottomrule
  \end{tabular}
\end{table}

\begin{table}[!htbp]
  \centering
  \caption{Giant Score: combining semantic content with structural position.}
  \label{tab:giant}
  \begin{tabular}{rlrrr}
    \toprule
    Rank & Scholar (dates)               & Score & Calc.\ Degree & Paths \\
    \midrule
     1   & J.\,F.\ Pfaff  (1765--1825)  & 0.841 & 0.638 & 33 \\
     2   & Leibniz        (1646--1716)  & 0.597 & 0.569 & 47 \\
     3   & J.\ Bernoulli  (1655--1705)  & 0.423 & 0.551 & 45 \\
    12   & Euler          (1707--1783)  & 0.244 & 0.529 & 45 \\
    17   & Gauss          (1777--1855)  & 0.167 & 0.531 & 29 \\
    91   & Newton         (1643--1727)  & 0.025 & 0.534 &  4 \\
    \bottomrule
  \end{tabular}
\end{table}

\subsection{The Republic of Letters as a multi-predicate institutional transition}

If the 17th-century watershed reflects the formation of a more durable
transmission infrastructure, the change should appear not only in path counts
but across independent attributes of scholarly activity. We compare pre- and
post-Leibniz windows within the induced subgraph using seven indicators drawn
from orthogonal predicate dimensions (Table~\ref{tab:rol}). The question is
whether a coherent cluster of fields, languages, employers, and voluntary
associations re-organizes around Leibniz without erasing the earlier layers of
the network.

The strongest single indicator is learned-society membership. Before the
Leibniz window, 6.5\% of scholars in the network held recorded society
memberships, averaging 0.08 per person; after Leibniz, the rate rose to 82.1\%,
averaging 3.81 per person---a $46\times$ increase. This is the most direct
network analogue of Mokyr's argument: voluntary scholarly associations became
part of the durable infrastructure through which knowledge circulated. The
three continuity indicators (field, language, employer) each measure how much
the composition of academic activity changed from the pre- to the post-period;
all three exceed the no-change baseline of $0.5$ ($0.614$, $0.603$, $0.603$),
indicating \emph{continuous transformation}---the cluster centered on calculus
developed and differentiated through Leibniz rather than replacing what came
before. That three independent dimensions exhibit this pattern simultaneously
shows the shift pervaded the whole social infrastructure of knowledge
production: not merely what was studied, but how, where, and in what language it
was communicated. The Student Top-Tier Production Rate of $1.000$---every one of
Leibniz's direct students generated descendants reaching the highest tier of
network influence---is the signature of a self-sustaining transmission
infrastructure. The joint movement of seven indicators across orthogonal
dimensions establishes a system-level institutional transition, not an artefact
of better modern documentation in any single dimension.

\begin{table}[!htbp]
  \centering
  \caption{Indicators of institutional transformation centered on Leibniz.}
  \label{tab:rol}
  \begin{tabular}{lll}
    \toprule
    Indicator & Value & Evidence \\
    \midrule
    Society explosion (Post/Pre)   & $0.08\!\to\!3.81$ ($46\times$) & Strong \\
    Field continuity               & $0.614$        & Strong \\
    Language continuity            & $0.603$        & Strong \\
    Employer continuity            & $0.603$        & Strong \\
    Inst.\ hub diversification     & $2.5\times$    & Moderate \\
    Student top-tier rate          & $100\%$        & Very strong \\
    Influence diffusion entropy    & $3.401$        & Very strong \\
    \bottomrule
  \end{tabular}
\end{table}

\subsection{Upstream convergence and the Monastery Wall}

We now turn upstream of the Leibniz watershed. Among the 213 individuals with
the maximum path count of 54 (surpassing even Leibniz), five Islamic scholars of
the 12th--13th centuries emerge as the furthest-upstream hubs of the entire
network (Table~\ref{tab:islamic}). The chain terminates with Grigorios
Choniades, a Byzantine scholar who studied in Persia and carried Islamic
astronomical and mathematical knowledge back to the Greek-speaking world.

That 54 of 64 Fields Medalists ($84.4\%$ at our reference frontier size
$\mathrm{FS}=20{,}000$) trace genealogies converging on five individuals
constituting less than $0.001\%$ of the database reveals a striking
path-dependency in documented human-capital transmission. Stated at this
magnitude the finding is not a counterfactual claim that ``Islamic science
caused Fields Medals''; it is the claim that, on the only relation we observe at
scale, a microscopic slice of the node set carries an overwhelming share of the
paths on which the tracer set depends. This is consistent with the qualitative
arguments of \citet{Saliba2007} and \citet{Itoh1993}, to which it adds a
quantitative structural dimension. The figure is a lower bound in two senses:
it is frontier-size-dependent, and Wikidata is biased toward the Western archive
and the doctoral-advisor idiom, so genuine debts to figures absent from the
graph can only dilute the apparent share of any chokepoint in future, richer
data.

The mentor--student predicate captures only one channel of transmission. The
12th-century Great Translation Movement---in which Arabic renderings of Greek
originals and original Arabic works were translated into Latin at Toledo,
Sicily, and northern Italy \citep{Itoh1993}---moved knowledge through books and
translators, not personal mentorship, and is by construction invisible to a
mentor--student genealogy. A further contribution of this paper is to show that
it is not invisible to \emph{semantic} analysis of the network's nodes. Using
the instrument's Unbinding operation to extract the field component from each
node's composite vector and tracking 50-year windows from 1100 to 1800, we
detect a coherent Toledo--Maragha astronomy signal during 1250--1300
($+0.0025$ relative to base) and a stronger Renaissance signal during 1450--1500
($+0.0065$, coinciding with Copernicus's lifetime), while algebra remains within
$\pm 0.001$ of the random baseline of $0.5$ across the same windows
(Table~\ref{tab:toledo}). The window construction, the Unbinding operation, and
the corroboration with Holy-Roman-Empire language-shift data are detailed in
Appendix~H.

Why does the chain terminate with the five Islamic scholars rather than
continuing? Despite exhaustive parameter exploration, no genealogy extends
beyond the 57th generation. Every path terminates at one of two figures who
predate the five by roughly a century: Archbishop Lanfranc of Canterbury
(1005--1089), who taught at the Abbey of Bec, or Ivo of Chartres (1040--1116);
their own mentors are unrecorded. We term this systematic cessation the
\emph{Monastery Wall}. As \citet{Rashdall2010} and \citet{DeRidder1992}
document, education in the 10th and 11th centuries revolved around
\emph{scholae monasticae} and cathedral schools, where the monastic ideal of
\emph{stabilitas loci} made education a communal affair oriented toward
scriptural understanding---channels that by their nature generated no structured
mentor--student records. The 11th-century shift from this regime to the personal
academic lineages of the cathedral schools and nascent universities is the
supply-side complement to the demand-side transition that \citet{Cantoni2014}
study.

The chronological profile of the 213 maximum-path scholars makes the boundary
visible at the population level: a single 11th-century entry (Lanfranc), only 13
individuals in the 12th century, a ramp through the scholastic 13th--15th
centuries, a peak at the 16th-century inception of the Scientific Revolution
($37.1\%$ of the 213), and a 17th-century tail after which Leibniz alone carries
the maximum. The single 11th-century entry is not a statistical anomaly but the
literal Monastery-Wall boundary in the data: the earliest scholar at whom the
personal-mentorship chain still attaches to a name on record. Where the
17th-century watershed was a shift in the social architecture of production
within an existing regime, the 11th-century transition was a shift in the medium
of transmission itself. The Islamic convergence and the Monastery Wall are, in
this sense, two faces of one institutional discontinuity: modern lineages pass
through these five scholars because, on the far side of them, ``personal
mentorship'' had not yet crystallized into a record-generating institution in
Europe; on the near side, it had.

\begin{table}[!htbp]
  \centering
  \caption{The five Islamic and Byzantine scholars at the headwaters of the
           Fields Medalist genealogy, in order of mentor-to-student succession.
           All five lie on the chain through which 54 of 64 Fields Medalists
           trace their genealogy.}
  \label{tab:islamic}
  \begin{tabular}{rlll}
    \toprule
    Order & Scholar                       & Dates; Birthplace          & Paths \\
    \midrule
    1 & Sharaf al-D\=in al-\d{T}\=us\=i  & 1135--1201; Tus (Iran)      & 54 \\
    2 & Kam\=al al-D\=in bin Y\=unis      & 1156--1242; Mosul          & 54 \\
    3 & Na\d{s}\=ir al-D\=in al-\d{T}\=us\=i & 1201--1274; Tus (Iran)  & 54 \\
    4 & Shams al-D\=in al-Bukh\=ar\=i     & 1254--1300; Bukhara        & 54 \\
    5 & Grigorios Choniades               & 1240--1320; Constantinople & 54 \\
    \bottomrule
  \end{tabular}
\end{table}

\begin{table}[!htbp]
  \centering
  \caption{Rate of HDC signal change between adjacent 50-year windows for
           algebra and astronomy, extracted via Unbinding of the field
           component from the composite hyperdimensional vector of all scholars
           active in each window. $\blacktriangle$ / $\triangledown$ mark notable
           increases / decreases. Astronomy departs upward in two windows aligned
           with the second Toledo translation period (1250--1300) and the
           Copernican Renaissance (1450--1500); algebra remains near the baseline
           of $0.5$.}
  \bigskip
  \label{tab:toledo}
  \small
  \begin{tabular}{lrrrr}
    \toprule
     & \multicolumn{2}{c}{Algebra} & \multicolumn{2}{c}{Astronomy} \\
    \cmidrule(lr){2-3}\cmidrule(lr){4-5}
    Window     & Signal & $\Delta$ & Signal & $\Delta$ \\
    \midrule
    1100--1150 & 0.4999 & (base)    & 0.5003 & (base) \\
    1150--1200 & 0.5001 & $+0.0001$ & 0.4999 & $-0.0003$ \\
    1200--1250 & 0.4998 & $-0.0003$ & 0.4999 & $\sim$0 \\
    1250--1300 & 0.4997 & $-0.0001$ & 0.5024 & $+0.0025\,\blacktriangle\blacktriangle$ \\
    1300--1350 & 0.5001 & $+0.0004$ & 0.5021 & $-0.0003$ \\
    1350--1400 & 0.5000 & $-0.0001$ & 0.4994 & $-0.0027\,\triangledown\triangledown$ \\
    1400--1450 & 0.4995 & $-0.0004$ & 0.4987 & $-0.0007$ \\
    1450--1500 & 0.4994 & $-0.0001$ & 0.5052 & $+0.0065\,\blacktriangle\blacktriangle\blacktriangle\blacktriangle\blacktriangle\blacktriangle$ \\
    1500--1550 & 0.5003 & $+0.0009$ & 0.5033 & $-0.0019\,\triangledown$ \\
    1550--1600 & 0.4996 & $-0.0007$ & 0.5039 & $+0.0006$ \\
    1600--1650 & 0.5001 & $+0.0005$ & 0.5076 & $+0.0037\,\blacktriangle\blacktriangle\blacktriangle$ \\
    1650--1700 & 0.5019 & $+0.0018\,\blacktriangle$ & 0.5053 & $-0.0023\,\triangledown\triangledown$ \\
    1700--1750 & 0.5006 & $-0.0013\,\triangledown$ & 0.5050 & $-0.0002$ \\
    1750--1800 & 0.5010 & $+0.0004$ & 0.5079 & $+0.0029\,\blacktriangle\blacktriangle$ \\
    \bottomrule
  \end{tabular}
\end{table}

\subsection{The instrument as a finding, and a robustness analysis}
\label{sec:robust}

The findings so far are about the historical network. An unanticipated finding
concerns the instrument, and proves on reflection to be also about the history.
The VaCoAl architecture includes a block-level collision-tolerance mechanism
(``Don't Care'') that suppresses block votes when two distinct items map to the
same address. In single-step retrieval this guards against false majorities;
under iterative multi-hop traversal it acquires a second-order effect we did not
anticipate.

\begin{figure}[ht]
  \centering
  \includegraphics[width=0.95\linewidth]{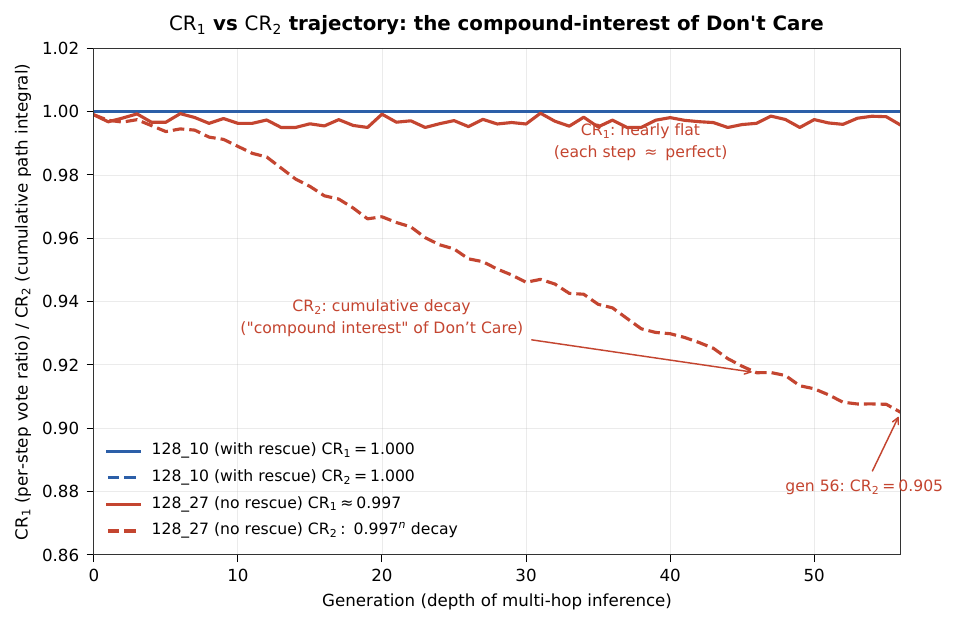}
  \caption{\textbf{Generational trajectories of $\mathrm{CR}_1$ and
    $\mathrm{CR}_2$.} PyVaCoAl in the $128\_27$ configuration (without rescue;
    main analysis) versus $128\_10$ (with rescue). In $128\_27$ the
    per-generation $\mathrm{CR}_1$ stays within $0.995$--$0.999$, so single-step
    retrieval is near-perfect, but $\mathrm{CR}_2$ decays monotonically,
    reaching $\approx 0.905$ at the 56th generation. In $128\_10$ with rescue,
    both are pinned at $1.0$, suppressing the emergent selection mechanism
    entirely.}
  \label{fig:cr1cr2}
\end{figure}

Per-step confidence $\mathrm{CR}_1$ at sufficient memory depth stabilizes
slightly below unity---in our experiments $\mathrm{CR}_1\!\approx\!0.997$.
Because the path-integral score accumulates as
\begin{equation}
\mathrm{CR}_2(n) \;=\; \mathrm{CR}_2(n-1)\,\cdot\,\mathrm{CR}_1(n-1),
\label{eq:cr2}
\end{equation}
the deviation compounds exponentially with generational distance,
$\mathrm{CR}_2(n)\!\approx\!\mathrm{CR}_1^{\,n}$, so that at $n=56$,
$\mathrm{CR}_2\!\approx\!0.997^{56}\!\approx\!0.846$. On 25.5 million real paths
through a 57-generation DAG we measure $\mathrm{CR}_2\!\approx\!0.905$ at
generation 56---of the same order as the closed-form prediction, with the
residual gap consistent with survivor bias along deep paths
(Fig.~\ref{fig:cr1cr2}).

\begin{figure}[!htbp]
  \centering
  \includegraphics[width=0.95\linewidth]{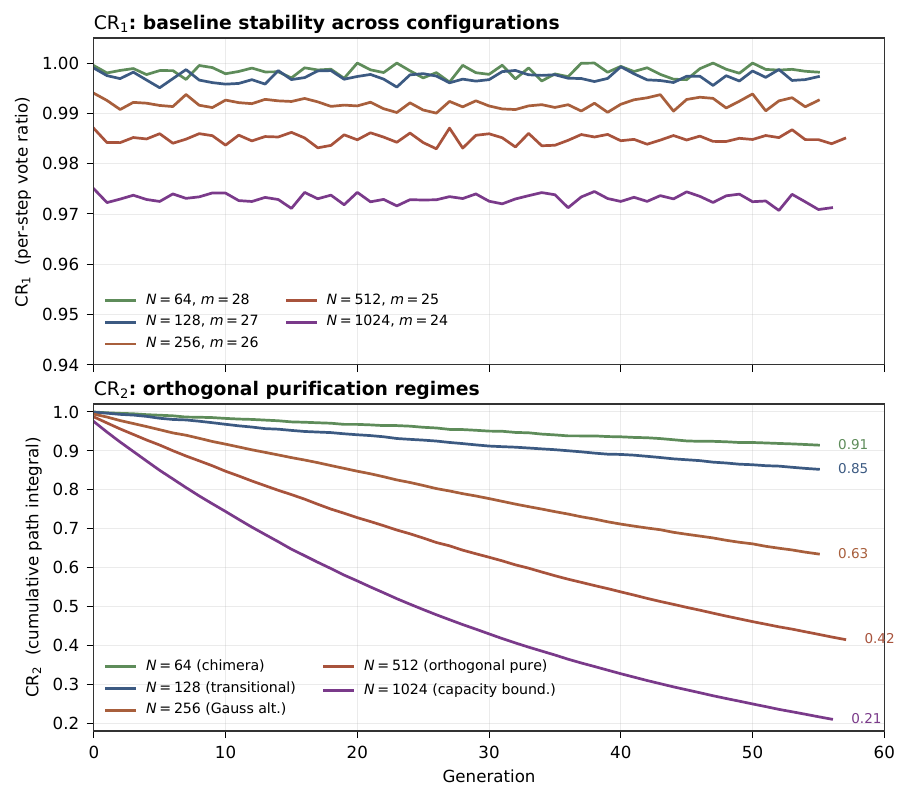}
  \caption{\textbf{$\mathrm{CR}_1$ and $\mathrm{CR}_2$ overlay across
    configurations.} $\mathrm{CR}_1$ (upper band) stays near $1.0$ regardless of
    dimensionality, confirming the macroscopic signal is never lost.
    $\mathrm{CR}_2$ (lower trajectories) decays at rates depending on $N$:
    low-$N$ configurations remain persistently high (the ``chimera'' regime);
    $N=512$ shows steep yet smooth decay (orthogonal purification) and reaches
    the maximum depth of 57 generations; $N=1024$ shows the steepest decay, the
    capacity boundary under $N\!\cdot\!2^{m}=2^{34}$.}
  \label{fig:overlay}
\end{figure}

The within-frontier ranking of candidates is therefore not arbitrary: short,
direct genealogical routes accumulate less penalty than circuitous, long ones,
and the instrument selects them accordingly---a \emph{structural Occam's razor}.
This emergent property is functionally analogous to spike-timing-dependent
plasticity in biological neural circuits \citep{BiPoo1998}: connections
reinforced by short, direct sequences strengthen while those traversed only
through long, circuitous chains weaken. We did not design this in; the
algebraic collision-tolerance mechanism, combined with the multiplicative
structure of $\mathrm{CR}_2$, generates it. The behavior is not a
single-configuration coincidence: sweeping the block count $N$ from 64 to 1024
under the fixed total-capacity constraint $N\!\cdot\!2^{m}=2^{34}$,
$\mathrm{CR}_1$ remains above $0.975$ throughout while $\mathrm{CR}_2$ traces a
family of decay curves encoding the trade-off between orthogonal resolution
(scaling with $N$) and per-block collision tolerance (scaling with $2^{m}$); the
single configuration that matches the maximum genealogical depth of 57
generations is $N=512$ (Fig.~\ref{fig:overlay}), a phase-transition signature
rather than a tuning artefact, and the strongest internal evidence that the
historical findings are not balanced on a particular parameter choice.

This finding is consequential for historical interpretation, and it is also the
part of the paper most likely to invite skepticism. A reasonable concern is
whether the extreme path-dependence we report---84\% of genealogies through five
individuals, the $10\!:\!1$ hourglass asymmetry---is a property of history or an
artefact of an instrument that itself generates path-dependence. Four points
respond, in decreasing order of abstraction.

\textit{Scale:} the instrument's path-dependence operates at
$\mathrm{O}(10^{-3})$ per-step deviations; the historical path-dependence
operates at $54/64$ Medalists and ratios of $10\!:\!1$ and $46\!:\!1$---not
quantities the instrument could conjure. \textit{Shape:} the instrument's decay
is smooth and monotonic along generational distance; the historical findings
exhibit discrete structural discontinuities (a wall at the 11th century, a
bottleneck at five 12th-century individuals, a watershed at 1700) with no
analogue in the instrument's internal dynamics. \textit{Switchability:} the
Rescue-mode configuration (which pins $\mathrm{CR}_1\!=\!1.0$ and eliminates the
emergent penalty entirely) produces the same Leibniz hourglass, the same
Republic-of-Letters indicators, the same Islamic convergence, the same Monastery
Wall; the historical findings are visible with or without the mechanism.
\textit{Direction:} the emergent mechanism acts as an Occam's razor on the
candidate ordering \emph{within} each frontier, not on the selection of the
frontier itself.

That the macro-structures survive switching off the razor (Rescue mode)
demonstrates that it refines rather than generates them. In Don't Care mode it
sharpens the within-frontier ranking so that short, direct, semantically
coherent lineages rise above deep, circuitous ones: under PyVaCoAl's
$\mathrm{CR}_2$-based ordering the calculus pioneers---Leibniz (2), Euler (7),
J.~Bernoulli (7), Lagrange (17), Poisson (20)---all enter the top 20, whereas
under a naive Python \texttt{dict} lexicographic ordering they are replaced by
anatomists, surgeons, physicians, and humanists whose vote counts happen to
survive the alphabetical cut, and Leibniz is demoted to rank 20
(Table~\ref{tab:votes}). The broader methodological moral is general: on real
data at a scale exceeding the writ of case-study investigation, path-dependence
can arise not only from historical contingency but from the measurement
architecture itself, and the constructive response is not to hope the
architecture is innocent but to characterize its bias in closed form---here, a
fourth-decimal-place per-step effect that the first-order historical findings
overwhelm.

\begin{table}[ht]
\centering
\caption{Top-20 vote comparison: PyVaCoAl ($N=128, m=27$, no rescue) versus
         Python \texttt{dict}. Votes in units of $10^{4}$.}
\label{tab:votes}
\scriptsize
\setlength{\tabcolsep}{2pt}
\begin{tabularx}{\columnwidth}{rXr @{\hspace{6pt}} rXr}
\toprule
\multicolumn{3}{c}{PyVaCoAl (no rescue)} & \multicolumn{3}{c}{Python \texttt{dict}} \\
\cmidrule(lr){1-3}\cmidrule(lr){4-6}
Rk & Scholar (d.) & Votes & Rk & Scholar (d.) & Votes \\
\midrule
 1 & J. Thomasius (1684)  & 1,510 &  1 & J. Thomasius (1684)         & 1,528 \\
 2 & Leibniz (1716)       & 1,321 &  2 & B. Meisner (1626)           & 1,378 \\
 3 & B. Meisner (1626)    & 1,268 &  3 & C.A. Hausen (1743)          & 1,339 \\
 4 & W. of Ockham (1349)  & 1,214 &  3 & A.G. K\"astner (1800)       & 1,339 \\
 5 & Duns Scotus (1308)   & 1,182 &  5 & A. Rhode (1633)             & 1,167 \\
 6 & A. Hegius (1498)     & 1,169 &  6 & A. Hegius (1498)            & 1,140 \\
 7 & Euler (1783)         & 1,054 &  7 & W. of Ockham (1349)         & 1,092 \\
 7 & J. Bernoulli (1748)  & 1,054 &  8 & Duns Scotus (1308)          & 1,062 \\
 9 & T. \`a Kempis (1471) & 1,032 &  9 & J. Dubois (1555)            & 1,060 \\
10 & G. Groote (1384)     & 1,015 & 10 & T. \`a Kempis (1471)        & 1,012 \\
11 & Gonsalvus (1313)     & 1,003 & 11 & G. Groote (1384)            &    992 \\
12 & P. Olivi (1298)      &    987 & 12 & Gonsalvus (1313)           &    878 \\
13 & C.A. Hausen (1743)   &    970 & 13 & P. Olivi (1298)            &    865 \\
14 & A.G. K\"astner (1800)&    968 & 14 & J. Peckham (1292)          &    840 \\
15 & A. Rhode (1633)      &    961 & 15 & J.W.v. Andernach (1574)    &    839 \\
16 & J. Peckham (1292)    &    957 & 16 & H. Fabricius (1619)        &    835 \\
17 & Lagrange (1813)      &    945 & 17 & G. Falloppio (1562)        &    832 \\
18 & J. Martini (1649)    &    944 & 18 & Bonaventure (1274)         &    816 \\
19 & Bonaventure (1274)   &    928 & 19 & J.L. d'\'Etaples (1536)    &    804 \\
20 & Poisson (1840)       &    903 & 20 & Leibniz (1716)             &    787 \\
\bottomrule
\end{tabularx}
\end{table}


\section{Discussion}

The network through which modern scientific knowledge was transmitted is not a
web but a sequence of structural transitions. The 17th-century hourglass at
Leibniz, with its $10\!:\!1$ asymmetry, quantifies the multiplier that
institutional integration applied to human-capital transmission; the 46-fold
expansion in learned-society memberships across the same window, with six
further independent indicators, supplies a structural anatomy of the
institutional formation \citet{Mokyr2016} identifies with the Republic of
Letters. Reframed as transmission rather than priority, the Newton--Leibniz
dispute admits a quantitative answer---Newton at rank 91, Leibniz at rank 2 on
the Giant Score---that operationalizes the distinction between the
\emph{possession} and the \emph{transmission} of upper-tail human capital.
Upstream, 84\% of lineages converge on five Islamic and Byzantine scholars who
are less than $0.001\%$ of the database, and the chain terminates at the
11th-century Monastery Wall, the more foundational of the two transitions.

For economists, the upshot is that ``institutions'' are not a residual category
to be invoked after a discovery story is complete. The hourglass, the
learned-society surge, and the multi-predicate continuity together depict a
durable network good whose architecture changed at identifiable centuries. We do
not estimate growth rates from the graph; the structural facts we report are a
precondition for any growth narrative that would treat the cost of transmitting
upper-tail human capital as endogenous, and they suggest concrete objects for
identified follow-up work---for example, the formation of academies as a
treatment whose downstream lineage effects are now, in principle, measurable.

Several limitations temper the findings. Wikidata's recording biases---against
non-Western, pre-university, and non-doctoral transmission---condition all
results, though they work \emph{against} our most striking upstream finding: a
fuller archive could only dilute the apparent share of any specific chokepoint.
We present no causal identification strategy; the findings are
descriptive-structural claims. The mentor--student predicate captures only one
mode of transmission, and although we recover a second (books and translations)
semantically, we have not mapped all modes. A different tracer set would change
absolute counts and might shift the identities of the chokepoint individuals,
but not the question of where the topology concentrates.

A word on responsible interpretation is in order, because macroscopic statistics
can be misread as verdicts on whole cultures. Our graphs attach numbers to
lineages \emph{as Wikidata currently encodes them}, not to moral worth,
civilizational ranking, or ancestry. A concentration of path traffic through a
small set of pre-modern names is a fact about a database under a specific
relation, and is best read as a symptom of who received documentary attention.
The appropriate use of the finding is to generate disciplined hypotheses for
historians, linguists, and area specialists. If a future release of Wikidata
alters a handful of high-degree medieval edges and our contours move
accordingly, that is a success of the programme, not a failure.


\section{Conclusion}

Working with all 64 historical Fields Medalists as a fixed, \emph{ex ante}
tracer set on the integrated Wikidata mentor--student graph, we have
reconstructed for the first time at this scale the 25.5-million-path directed
acyclic graph through which modern mathematical lineages connect to their
pre-modern antecedents, and extracted three structural facts about the
transmission of upper-tail human capital. They cohere into one picture. Across
roughly nine centuries of documented mentorship, the social architecture of
transmission did not evolve as a smooth gradient but moved through identifiable
transitions: the 11th-century Monastery Wall, the prior condition at which
personal academic lineage first becomes record-generating in Europe; and the
17th-century Leibnizian hourglass, through which an already-institutionalized
regime acquired a $10\!:\!1$ amplification, a 46-fold expansion in
learned-society membership, and a coherent re-organization across seven
predicate dimensions. Between Wall and hourglass, the highest-traffic
individuals trace a profile that peaks in the 16th-century inception of the
Scientific Revolution and yields to Leibniz alone thereafter. The
Newton--Leibniz dispute, reframed under this geometry, resolves in favor of
transmission over possession.

The instrument is a second contribution, but a subordinate one: a deterministic,
algebraic graph-traversal method that makes exhaustive structural measurement on
civilization-scale relational data feasible, and whose measurement bias is
characterized in closed form rather than assumed away. We hope the larger
methodological point outlasts the particular findings: that the economic history
of knowledge now admits a structural, exhaustive mode of measurement, and that
the facts it yields are the natural raw material for the identified, causal work
the field does best.


\section*{CRediT authorship contribution statement}
\textbf{Hiroyuki Chuma:} Conceptualization, Methodology, Software,
Formal analysis, Investigation, Data curation, Writing -- original draft,
Writing -- review \& editing. \textbf{Kanji Otsuka:} Methodology
(instrument architecture), Writing -- review \& editing.
\textbf{Yoichi Sato:} Methodology (instrument architecture),
Writing -- review \& editing.

\section*{Declaration of competing interest}
The authors declare that they have no known competing financial interests or
personal relationships that could have appeared to influence the work reported
in this paper.

\section*{Funding}
This research received no specific grant from any funding agency in the public,
commercial, or not-for-profit sectors.

\section*{Acknowledgments}
We thank the Wikidata, Mathematics Genealogy Project, and MacTutor History of Mathematics Archive communities for maintaining the data infrastructure upon which this study rests. 
We used Anthropic's Claude AI assistant for Japanese-to-English translation and editorial assistance in the preparation of this manuscript. Additionally, Claude Code and the Cursor AI development environment were utilized to build our Python application, "PyVaCoAl." 
The authors retain full responsibility for all scientific claims, theorems, and proofs presented in this work.

\section*{Data availability}
The underlying data are drawn from Wikidata, which is publicly available under a
CC0 license. The Wikidata extraction queries (SPARQL), the genealogy-search
configurations, the reference-vector specifications, the complete 64-Medalist
tracer set, the PyVaCoAl reference implementation, and all tabular results
reported here are deposited in a public repository (DOI to be assigned upon
acceptance) and are available to editors and referees on request during peer
review. All results are produced by a deterministic algorithm with no random
seeds, so the reported figures are exactly reproducible from the deposited code
and inputs.

\appendix

\titleformat{\section}[block]
  {\large\bfseries\sffamily}
  {Appendix \thesection.}
  {0.8em}
  {}
\titlespacing*{\section}{0pt}{2.5ex plus 1ex minus .2ex}{1.5ex plus .2ex}

\bigskip
\noindent\textit{Appendix.}\ The appendices document the data construction
(A--B), the deterministic graph-traversal instrument (C--G), the proofs (I), and
the robustness checks and reproducibility detail (J--K) underlying the results
in the body. Appendices~C--G concern the measurement instrument; a reader
interested only in the economic-history findings may treat them as supporting
material.
\bigskip

\section{Data}

From Wikidata's SPARQL endpoint (snapshot accessed 2026), we extracted all
entities linked by the \emph{doctoral advisor} (P184), \emph{student}
(P802), and \emph{student of} (P1066) predicates within the network
reachable from mathematics-oriented people.  The raw dataset contained
approximately $470{,}000$ records.  A pre-purification pass detected and
removed $369$ bidirectional links (cycles), yielding a clean directed
acyclic graph (DAG) of $372{,}853$ scholars.  Wikidata's systematic
biases---over-representation of English-, German-, and French-language
scholarship; denser modern than medieval records; better coverage of
university-affiliated than court or non-Western practitioners---work
against our Islamic convergence finding; the $84\%$ rate is therefore a
lower bound.

\section{Genealogical search on a combinatorially explosive DAG}

Starting from each of the 64 Fields Medalists, we traced mentor chains
backward iteratively.  Academic genealogy on the Wikidata mentor--student
graph is not a tree: a scholar may have multiple recorded doctoral
advisors; any given mentor typically has many students; and common
ancestors propagate through overlapping descendant lineages.  The
resulting structure is a DAG in the strict mathematical sense, and the
number of distinct paths from a source node to its ancestors grows
multiplicatively with generational depth even when the number of distinct
nodes is modestly bounded.  Starting from 64 Fields Medalists on a DAG of
$372{,}853$ nodes, the path count reaches $25.5\!\times\!10^{6}$ at depth
$57$---a path-to-node ratio of approximately $68\!:\!1$.  A naive
exhaustive search explodes well before this depth, so the search is
executable only under a finite Frontier Size (FS) cap, which we set to
$20{,}000$.  Every genealogical result in this paper is therefore
FS-relative.

\textit{Failure of standard lookup primitives.}  We verified the failure
of three standard alternatives under the data-scale conditions of our
experiment.  (i)~Python hash tables: at $\mathrm{FS}=25{,}000$, the Python
\texttt{dict}-based baseline exhausted $128$~GB of main memory and became
unexecutable.  When \texttt{dict} runs at smaller FS, it ranks the
within-FS candidates by lexicographic order of their identifiers, an
ordering with no historical significance.  (ii)~GPU-based HDC libraries:
the standard open-source framework \texttt{torchhd}~\citeyearpar{Heddes2023}
requested over $100$~GB of VRAM for the semantic-analysis stage and
terminated with out-of-memory error on an NVIDIA RTX 3060 ($12$~GB VRAM);
in GPU fallback mode, PCIe-bus bandwidth collapses the in-memory benefit
entirely.  (iii)~Graph-theoretic centrality measures: PageRank and
betweenness centrality complete in reasonable time on the full graph but
are field-blind---they cannot distinguish a mathematician from an
anatomist occupying the same topological position.  None of these
constraints reflects a poor tool choice; each is a direct consequence of
the huge DAG structure of deep genealogical data interacting with the
architectural assumptions of the standard toolkit (absolute exact match,
decorrelated hashing, dense floating-point matrix operations).  The
methodological question for any instrument operating under the irreducible
FS relativity imposed by the data is therefore: how are the within-FS
candidates ranked, and does the ranking carry semantic content?

\section{VaCoAl/PyVaCoAl architecture}

We constructed an algebro-deterministic HDC architecture based on
Galois-field algebra~\citeyearpar{Chuma2026Beyond} that replaces Kanerva's
probabilistic random projections~\citeyearpar{Kanerva1988} with deterministic
diffusion via Linear Feedback Shift Registers.  For each of the
approximately $1{,}043$ hub individuals identified from the genealogy
traffic, we attached up to 12 predicate dimensions via Wikidata SPARQL
queries (academic field, language used, employer institution, society
membership, era, birthplace, and others).  Attributes were encoded as
million-dimensional binary hypervectors and subjected to Binding, Unbinding,
and Bundling operations, enabling continuous measurement of each node's
affinity to specific intellectual traditions via Hamming-distance
similarity.  The entire computation completed in 20--30 minutes on a
decade-old Intel Xeon E5-1650 v3 (2014) with no GPU.

\textit{Galois-field diffusion.}  Input data $P(x)$, represented as a
binary polynomial over GF(2) with coefficients $a_i\in\{0,1\}$, is mapped
to a high-dimensional vector via
\begin{equation}
F(x) \;=\; x^{m}P(x) \;+\; \bigl(x^{m}P(x) \bmod G(x)\bigr),
\label{eq:GFmap}
\end{equation}
where $G(x)$ is a degree-$m$ generator polynomial drawn from the class of
long-period polynomials over GF(2).  The choice of a finite field rather
than an unbounded ring is structural: the $\bmod\,G(x)$ operation ensures
that the transformation cycles within a bounded address space, and it is
precisely this cycling within a closed world that scatters similar inputs
into mutually unrelated directions, creating orthogonality
\emph{algebro-deterministically}.  The operation is implementable as a
single LFSR pass whose hardware cost is negligible.  Under the
primitive-polynomial conditions detailed in~\citeyearpar{Chuma2026Beyond}, a
single-bit input change triggers the \emph{avalanche effect}: approximately
$50\%$ of output bits flip.

\textit{Per-block diffusion.}  The input hyper-vector is partitioned into
$N$ segments of length $q=L/N$ (main analysis uses $N=128$,
$q\!\approx\!8000$, $L\!\approx\!10^{6}$).  Each block $b$ applies its own
independent LFSR diffusion with its own feedback polynomial $G_b(x)$ and
its own seed, producing an $m$-bit address.  This per-block architecture
is structurally necessary: under a hypothetical global diffusion, a
single-bit perturbation of the input would, by the avalanche effect,
propagate to all $N$ block addresses, corrupting every vote.  Under the
per-block diffusion actually implemented, such a perturbation affects only
the one or two blocks whose segments contain the flipped bit; the
remaining $N-O(1)$ blocks continue to output the correct entry address,
and the avalanche effect within the affected blocks diffuses their
erroneous votes uniformly across the address space.

\textit{Block division and majority voting.}  Each of the $N$ blocks
(64--1{,}024 in our experiments) functions as an independent SRAM/DRAM
lookup table with $2^{m}$ entries.  In the write phase, a unique Entry
Address (EA) is written to the addressed location in every block.  In the
read phase, each block $i$ votes for an EA $v_i$:
\begin{equation}
W_{\text{VaCoAl}} \;=\; \arg\max_{v}\; \sum_{i=1}^{N}\delta(v_i, v),
\label{eq:majorityvote}
\end{equation}
where $\delta$ is the Kronecker delta.  Erroneous votes scatter uniformly
across the address space (a flat field), while correct votes concentrate
on a single peak.

\textit{Collision tolerance and the rescue circuit.}  When different items
map to the same block address, VaCoAl marks the block as \emph{Don't Care}
and excludes its vote from the aggregation in that retrieval step.  Under
multi-hop reasoning, this acquires a second-order property whose
closed-form magnitude appears in Appendix~F.  In the complementary
\emph{Rescue} mode (rescue rate $\mathrm{RR}>0$), a four-stage pipeline
explicitly intercepts and resolves collisions: sample accumulation at
write time; per-block address-sorted finalisation; binary search at query
time against per-block auxiliary arrays; and segment exact match within
each candidate interval.  With $\mathrm{RR}=1$ this pipeline pins
$\mathrm{CR}_1\!\equiv\!1.0$ and produces output bitwise identical to the
Python \texttt{dict} baseline (verified across all $25.5\!\times\!10^{6}$
records).  Main-text analysis uses $\mathrm{RR}=0$.

\textit{HDC algebraic operations.}  Three operations are central to the
semantic analysis: \textit{Binding} ($A\otimes B$) via XOR and cyclic shift
over GF(2), at cost $O(N)$; \textit{Bundling} ($A+B$) via component-wise
majority vote; and \textit{Unbinding} as the inverse of Binding, exploiting
the algebraic reversibility of XOR over GF(2) for lossless concept
decomposition.  Similarity is measured by Hamming distance; for
quasi-orthogonal vectors in an $L$-dimensional binary space, the expected
similarity between uncorrelated vectors is $0.5$.

Equation~\eqref{eq:GFmap} is formally identical to the systematic-code
form of a BCH code, yet the purpose is diametrically opposite.  BCH codes
expend elaborate computation (syndrome decoding) to force convergence to a
unique correct codeword.  VaCoAl abandons this convergence logic and uses
Galois-field operations as a \emph{scoreboard} for relative similarity and
path quality.  On massive DAGs subject to combinatorial explosion, an
absolute exact match cannot in principle exist; VaCoAl embeds a cognitive
bound---the Frontier Size---into the architecture and ranks within-FS
candidates by the path-integral score $\mathrm{CR}_2$.

\section{Confidence scores}

The architecture produces two confidence scores: $\mathrm{CR}_1$ is the
block-majority-voting rate at a single retrieval step;
$\mathrm{CR}_2(n) = \mathrm{CR}_2(n\!-\!1)\cdot\mathrm{CR}_1(n\!-\!1)$ is
the generational product, a path integral that accumulates confidence
multiplicatively along a chain.  In Rescue mode, $\mathrm{CR}_1\equiv 1.0$
identically.  In Don't Care mode with the $(N=128,\,m=27)$ configuration,
$\mathrm{CR}_1\!\approx\!0.995$--$0.999$; the residual per-block collisions
contribute the minute analog variance.  Within-FS candidate ranking is by
$\mathrm{CR}_2$; hash-based baselines lack any such score and fall back
to lexicographically vacuous ordering.  At sufficient memory depth
$\mathrm{CR}_1\!\approx\!0.997$, yielding the closed-form prediction
$\mathrm{CR}_2(n)\!\approx\!0.997^{n}$ that matches measured values across
all $25.5\!\times\!10^{6}$ paths.

\section{Mathematical isomorphism with biological neurons}

The output $y_{\text{bio}}$ of a biological neuron is the application of
a nonlinear activation $\varphi$ to the inner product of input
$\mathbf{x}\in\mathbb{R}^{D}$ with synaptic weights $\mathbf{w}\in\mathbb{R}^{D}$:
$y_{\text{bio}} = \varphi(\mathbf{w}\cdot\mathbf{x} - \theta).$
For HDC bipolar vectors $\{-1,+1\}^{D}$, the inner product relates linearly
to Hamming distance: $\mathbf{w}\cdot\mathbf{x} = D - 2\,d_H(\mathbf{w},\mathbf{x}).$
In VaCoAl, with $\delta_i\in\{0,1\}$ the match decision for block $i$,
$y_{\text{vac}} = \mathbb{I}\!\left(\sum_{i=1}^{N}\delta_i > \theta_{\text{digital}}\right).$
By the Law of Large Numbers and the uniform error distribution guaranteed
by Galois-field diffusion, the vote count is a strictly monotone increasing
estimator of $\mathbf{w}\cdot\mathbf{x}$, yielding
$y_{\text{vac}}\equiv \mathbb{I}(\mathbf{w}\cdot\mathbf{x}>\theta_{\text{analog}})$---formally
identical to the biological rule.  Mapping PyVaCoAl's processing onto the
three-layer structure of a pyramidal neuron yields a structural flow
equivalence (dendrites $\to$ soma $\to$ axon corresponds to
per-block exact match $\to$ majority voting on $\mathrm{CR}_1$ $\to$
$\mathrm{CR}_2$ propagation), not a strict dynamical isomorphism: dendrites
exhibit graded NMDA nonlinearities and somatic integration is
distance-weighted, while VaCoAl's checking is discrete and its voting
equal-weight.  We further note a biological homology in the substitution
of a fixed algebraic logic circuit for stochastic random projection:
Fiete~et~al.~\citeyearpar{Fiete2008} and Chandra~et~al.~\citeyearpar{Chandra2025}
show that the Residue Number System code formed by grid cells in the
entorhinal cortex is projected onto hippocampal CA3 as a fixed random
projection, completed during early childhood; subsequent memory processes
execute as orthogonalising mappings onto the invariant vessel.  VaCoAl's
LFSR-on-SRAM/DRAM realisation of orthogonalisation is, in this sense, a
digital-logic reproduction of this biological finding---an instance of
what we call \emph{convergent computational equivalence}.  Compared with
Eliasmith's Semantic Pointer Architecture~\citeyearpar{Eliasmith2013}, which
solves the Binding Problem~\citeyearpar{Thagard2019} via circular convolution
at $O(N\log N)$ FFT cost with floating-point arithmetic, VaCoAl provides
an algebraically reversible solution via Galois-field XOR and shift at
$O(N)$ bitwise cost at single-bit precision.

\section{Emergent semantic selection: multi-configuration phase transitions}

Under a fixed total-capacity constraint $N\!\cdot\!2^{m}=2^{34}$
(corresponding to $64$~GB DRAM per the $(128,27)$ realisation), the
system does not simply degrade monotonically as the per-block collision
rate rises; it undergoes four macroscopic topological phase transitions,
each preserving a distinct, historically valid lineage structure
(Table~\ref{tab:phase}).

\textit{Chimera state ($N=64$).}  In the lowest-dimensionality regime
($N$-dim $=6{,}400$), the system lacks the orthogonal resolution to
separate massive overlapping subgraphs.  Traversal produces a fragmented
superposition that fuses German lineage elements (Hausen, K\"astner) with
analytical elements (Thomasius) while losing the central hub Leibniz
entirely.

\textit{Transitional emergence ($N=128$, main analysis).}  Orthogonal
resolution at $N$-dim $=12{,}800$ suffices to identify the primary
macroscopic structure.  The Leibniz lineage emerges strongly (Leibniz,
Euler, J.~Bernoulli), but residual Gauss-lineage crossover nodes (Hausen,
K\"astner) still persist within FS due to incomplete culling.

\textit{Bifurcation and alternate attractor ($N=256$).}  The specific
analog-noise profile at $0.282\%$ collision rate alters the $\mathrm{CR}_2$
fitness landscape.  At critical historical bifurcations, this
configuration marginally favours the alternative subgraph: the Leibniz
lineage is culled, and the lineage of Carl Friedrich Gauss (Pfaff, Hausen,
K\"astner) emerges instead---an alternate, historically accurate
macroscopic structure.

\textit{Orthogonal purification ($N=512$).}  At $N$-dim $=51{,}200$ and
collision rate $0.559\%$, robust orthogonality subdues the interference
entirely.  The system converges definitively on the Leibniz--Euler
lineage and purifies the extraction down to Lagrange and Poisson.  This
is the only VaCoAl configuration matching the maximum genealogical depth
of $57$ generations attained by the \texttt{dict} baseline at
$\mathrm{FS}=20{,}000$.

\textit{Capacity boundary ($N=1024$).}  Extending $N$ to $1024$ halves $m$
from $25$ to $24$ and doubles the count-based collision rate to $1.108\%$.
The Leibniz-dominant regime is retained, but the reachable depth drops to
$56$.

\begin{table}[!htbp]
\centering
\caption{Collision-rate analysis under fixed total capacity
         $N\!\cdot\!2^{m}=2^{34}$, and the empirical regime it produces.}
\label{tab:phase}
\small
\begin{tabular}{lrrrl}
\toprule
Configuration & Loc.\ rate & Count rate & Depth & Regime \\
\midrule
$N=64,\,m=28$    & $0.000103\%$ & $0.074\%$  & 55 & Chimera \\
$N=128,\,m=27$   & $0.000399\%$ & $0.144\%$  & 55 & Leibniz dominant (main) \\
$N=256,\,m=26$   & $0.001565\%$ & $0.282\%$  & 55 & Gauss dominant \\
$N=512,\,m=25$   & $0.006189\%$ & $0.559\%$  & 57 & Orthogonal purification \\
$N=1024,\,m=24$  & $0.024437\%$ & $1.108\%$  & 56 & Capacity boundary \\
\bottomrule
\end{tabular}
\end{table}

Three conclusions follow.  (i)~High collision rates are not the enemy; the
quadrupling of the count-based collision rate from $N=128$ to $N=512$
\emph{improves} rather than degrades semantic purity.  (ii)~The
Don't-Care-driven STDP-like decay is not a blunt instrument; a deep path
with coherent $\mathrm{CR}_2$ mass along a truly continuous academic
lineage survives, while a similarly deep path traversing ancient junctions
without semantic continuity is eliminated.  (iii)~The optimal balance
under a fixed-capacity budget is a trade-off between orthogonal resolution
(scaling with $N$) and per-block collision tolerance (scaling with
$2^{m}$)---not maximising either factor in isolation.

\textit{Speed and the FS phase transition.}  A phase transition occurs at
$\mathrm{FS}\!\approx\!20{,}000$: \texttt{dict} is faster below this
threshold, PyVaCoAl overtakes above it; at $\mathrm{FS}=25{,}000$,
\texttt{dict} exhausts $128$~GB and fails while PyVaCoAl completes stably.
On the HDC semantic-analysis stage, the CPU-only \texttt{torchhd} baseline
required $\sim\!60$~min ($2\times$ PyVaCoAl's $\sim\!30$~min); the GPU
variant terminated with OOM after requesting $>\!100$~GB of VRAM.

\section{Functional equivalence to STDP and CA3 recurrent collaterals}

Spike-Timing-Dependent Plasticity~\citeyearpar{BiPoo1998} in biological neural
circuits is often described as a physical mechanism of analog synaptic
dynamics.  Abstracting away the implementation substrate, the
computational problem it solves is: \emph{in multi-stage information
transmission, exponentially decay the weight of a path according to its
distance from the starting point, selectively preserving close, direct
paths.}  Under this abstract specification, the STDP function does not
inherently depend on millisecond-precision analog spikes but on the
computational structure of distributed representation and threshold
judgement; implementations are therefore not limited to analog circuits.
The most important characteristic of CA3 recurrent-collateral integration
emphasised by Rolls~\citeyearpar{Rolls2023} is the recursion by which past
computational results are fed back into current pattern completion.  The
update rule in PyVaCoAl is formally isomorphic: the cumulative confidence
of the previous generation, $\mathrm{CR}_2(n\!-\!1)$, is recursively fed
into the computation for the next generation, and only candidates
surviving $\mathrm{CR}_1$, $\mathrm{CR}_2$, and the FS cap become search
starting points for the next generation.  Three functional parallels with
CA3 obtain: full-pattern completion from partial patterns; natural
suppression of sub-threshold patterns (halting at $\mathrm{CR}_2<0.10$);
and winner selection among competing patterns (Don't Care defers premature
commitment).  This is convergent computational equivalence, not mimicry:
different implementation substrates arrived at the same functional
structure by independently solving the same computational problem.  The
digital algebraic implementation acquires three properties unachievable in
principle by analog neuromorphic implementations:
\textit{perfect reversibility} (Binding/Unbinding via GF(2) XOR and shift
is algebraically reversible); \textit{mathematical auditability} (the
$\mathrm{CR}_2$ path integral permits algebraic backtracking of why a
given answer was produced); and \textit{deterministic zero-collision
guarantee} in Rescue mode (verified empirically across all
$25.5\!\times\!10^{6}$ records).

\section{HDC signal trace of the Great Translation Movement}

The mentor--student predicate captures only one channel of knowledge
transmission.  The Great Translation Movement of the 12th century
transmitted knowledge through books and translators, not personal
mentorship, and is therefore by construction invisible to a
mentor--student genealogy.  We recover this transmission channel
indirectly through HDC semantic analysis: for each 50-year window $W_k$,
we Bundle the composite $H(W_k)$ over all scholars active during the
window (birth-year overlap) and Unbind the FIELD component.  The signal
at window $k$ for field $F\in\{\text{algebra},\text{astronomy}\}$ is
\begin{equation}
\sigma_F(k) \;=\; \frac{1}{L}\sum_{i=1}^{L}\mathbb{I}\!\bigl(
  \mathrm{Unbind}(H(W_k),\,\mathrm{FIELD})_i = \mathrm{Ref}_{F,i}\bigr),
\end{equation}
where $\mathrm{Ref}_F$ is the field-specific reference vector.  The
astronomy signal surges during 1250--1300 ($+0.0025$, coinciding with the
second Toledo translation period and the peak of the Maragha observatory
school) and again during 1450--1500 ($+0.0065$, the Renaissance peak
overlapping with Copernicus's lifetime).  Algebra shows no comparable
surge.  Copernicus's documented borrowing of the mathematical models of
al-Urd\=i (1200--1266), al-\d{T}\=us\=i (1201--1274), and
Ibn al-Sh\=a\d{t}ir (1304--1375)~\citeyearpar{Saliba2007}---composed 200--300
years before his time and translated from Arabic into Latin via
intermediaries---is consistent with the semantic trace detected at
1250--1300 and its Renaissance amplification.  The asymmetry between
algebra and astronomy is therefore a substantive observation rather than
a by-product of window width: a flat discipline would wobble near the
random $0.5$ line, whereas astronomy repeatedly departs in the same
direction whenever translation episodes and Copernican-era activity
plausibly raised the public salience of instruments, tables, and
geometrical models.  The exercise does not prove that any individual
Medalist's intellectual DNA passed through a Toledo manuscript; it shows
that, at population level, the semantic field vector tilts in a way that
a purely topological graph test could never register.

\section{Proof: false-positive probability vanishes asymptotically}

Galois-field diffusion distributes random noise uniformly across the
address space of size $M=2^{m}-1$.  The per-block false-positive rate is
$p=1/M$.  Per-block diffusion ensures that a single-bit perturbation
affects only $O(1)$ blocks whose segments contain the flipped bit; the
remaining $N-O(1)$ blocks output the correct Entry Address.  Within the
affected blocks, the avalanche effect diffuses erroneous votes uniformly,
producing the flat-field noise distribution on which the Chernoff bound
relies.  The number of accidental coincident votes $X$ follows
$\mathrm{Binomial}(N,p)$.  In the worst case $N=M=1{,}000$: mean
$\mu = Np = 1$, majority threshold $\theta=500$, deviation $\delta=499$.
By the multiplicative Chernoff bound,
\begin{equation}
P_{\text{error}} \;=\; P(X\ge\theta)
\;\le\; \left(\frac{e^{\delta}}{(1+\delta)^{1+\delta}}\right)^{\mu},
\end{equation}
yielding $\ln P_{\text{error}} \approx 499 - 500\ln(500) \approx -2606$ and
$P_{\text{error}}\le e^{-2606}\approx 0$.  This bound is more than $2{,}500$
orders of magnitude smaller than the probability of randomly selecting a
specific atom from the observable universe ($\approx e^{184}$).
Noise-induced majority false positives in VaCoAl are, for all practical
purposes, impossible: the gap between noise floor ($\mu=1$) and majority
threshold ($\theta=500$) is insurmountable by probabilistic fluctuations
alone.

\section{Robustness checks}

\textit{Frontier Size sensitivity.}  Results are qualitatively stable
across $\mathrm{FS}\in\{10{,}000, 15{,}000, 20{,}000\}$.  At
$\mathrm{FS}=25{,}000$ the Python \texttt{dict} baseline fails (OOM), but
PyVaCoAl still produces stable figures.  The $84\%$ Islamic convergence
reported in the main text ranges from $\sim\!72\%$ at $\mathrm{FS}=5{,}000$
to $84\%$ at $\mathrm{FS}\ge 10{,}000$, plateauing thereafter.

\textit{Rescue Rate sensitivity.}  Between $\mathrm{RR}=0$ (pure Don't
Care; main analysis) and $\mathrm{RR}=1$ (full Rescue, eliminating the
$\mathrm{CR}_2$ penalty), all main-text historical findings are preserved
nearly bit-identically: the $84\%$ Islamic convergence, the Leibniz
$10\!:\!1$ hourglass, the Monastery Wall, the $46$-fold society-membership
explosion, the Newton rank of $91$ on the Giant Score.  The difference
between the two modes appears especially in the within-FS ranking of
sub-threshold candidates, precisely where $\mathrm{CR}_2$ adds semantic
content (Table~\ref{tab:votes}).

\textit{Memory-depth sensitivity and reachable depth.}  PyVaCoAl is not a
static hash-matching algorithm but a dynamic topological scanner.  Across
all configurations from $N=64$ ($N$-dim $=6{,}400$) to $N=1024$ ($N$-dim
$=102{,}400$), $\mathrm{CR}_1$ remains overwhelmingly stable above $0.975$,
confirming that the system never loses the core macroscopic signal despite
the intentional compression of memory depth.  The $\mathrm{CR}_2$
trajectories reveal the definitive signature of dimensional filtering:
$N=64$ retains persistently high $\mathrm{CR}_2$ (chimera state, lacking
orthogonal resolution); $N=512$ undergoes a steep yet smooth decay (the
mathematical manifestation of orthogonal purification, in which the
high-dimensional space actively filters out residual interference and
isolates the true analytical trajectory); $N=1024$ shows the steepest
decay, reaching the capacity boundary.  Viewed across all configurations,
the reachable depth is non-monotonic in $N$: $55$ generations for
$N=64,128,256$; $57$ for $N=512$; and $56$ for $N=1024$.  $N=512$ is the
only configuration matching the maximum depth of $57$ generations
attained by Python \texttt{dict} at $\mathrm{FS}=20{,}000$.  This
non-monotonicity is shown explicitly in the main-text Fig.~\ref{fig:overlay}
overlay and the per-generation trajectories of Fig.~\ref{fig:cr1cr2}.

\textit{Alternative tracer sets.}  Although a full head-to-head replication
with a different tracer set (e.g., Nobelists in physics) is beyond the
scope of the present paper, the structural logic of the exercise---that
a fixed, ex ante tracer set induces a well-defined subgraph on a DAG too
large for unaided intuition---does not depend on the particular choice of
Fields Medal.  The Wikidata predicates (P184, P802, P1066), the DAG
structure, the combinatorial explosion, and the FS relativity apply
identically to any modern elite science tracer set.  A different tracer
set would yield different absolute chokepoint identities; the
methodological architecture reported here would be unchanged.

\section{Reproducibility}

The complete algorithmic specification of VaCoAl/PyVaCoAl---including the
primitive-polynomial construction of the Galois-field diffusion, the
per-block LFSR addressing, the four-stage Rescue pipeline, the HDC
Binding/Unbinding/Bundling operations at single-bit precision, and the
$\mathrm{CR}_1$/$\mathrm{CR}_2$ computation---appears in the companion
preprint~\citeyearpar{Chuma2026Beyond} at a level of detail sufficient for
independent reimplementation.  Wikidata extraction queries (SPARQL),
genealogy-search configurations, HDC reference-vector specifications, the
complete 64-Medalist tracer set, and all tabular results of the present
paper are released as Supplementary data on a public repository (DOI to
be assigned at acceptance) and made available to editors and reviewers
upon request during peer review.  The combination of algorithmic
disclosure, public input data, deterministic algorithm (no random seeds),
and explicit closed-form bias characterisation satisfies the
reproducibility standard appropriate to a paper whose findings are
structural rather than statistical-inferential.


\end{document}